# Few-Shot Learning of Visual Compositional Concepts through Probabilistic Schema Induction

*Andrew Jun Lee[1] (andrewlee0@g.ucla.edu)    *Taylor Webb[3] (taylor.w.webb@gmail.com)
Trevor Bihl[4] (trevor.bihl@gmail.com)
Keith J. Holyoak[1] (kholyoak@g.ucla.edu)    Hongjing Lu[1,2] (hongjing@ucla.edu)

[1]Department of Psychology, [2]Department of Statistics, University of California, Los Angeles
[3]Microsoft Research, NYC, [4]Air Force Research Laboratory
*equal contribution

**Abstract**

The ability to learn new visual concepts from limited examples is a hallmark of human cognition. While traditional category learning models represent each example as an unstructured feature vector, compositional concept learning is thought to depend on (1) structured representations of examples (e.g., directed graphs consisting of objects and their relations) and (2) the identification of shared relational structure across examples through analogical mapping. Here, we introduce Probabilistic Schema Induction (PSI), a prototype model that employs deep learning to perform analogical mapping over structured representations of only a handful of examples, forming a compositional concept called a schema. In doing so, PSI relies on a novel conception of similarity that weighs object-level similarity and relational similarity, as well as a mechanism for amplifying relations relevant to classification, analogous to selective attention parameters in traditional models. We show that PSI produces human-like learning performance and outperforms two controls: a prototype model that uses unstructured feature vectors extracted from a deep learning model, and a variant of PSI with weaker structured representations. Notably, we find that PSI's human-like performance is driven by an adaptive strategy that increases relational similarity over object-level similarity and upweights the contribution of relations that distinguish classes. These findings suggest that structured representations and analogical mapping are critical to modeling rapid human-like learning of compositional visual concepts, and demonstrate how deep learning can be leveraged to create psychological models.

# Introduction

A fundamental aspect of human intelligence is few-shot learning: the ability to grasp concepts from just a few examples. In a visual few-shot learning task, learners are given a small set of class-labeled images (e.g., "image 1 belongs to category A") and must correctly classify a new unlabeled *target* image. A decade ago, deep learning models struggled with few-shot learning, even for perceptual categories defined by simple visual features (Lake et al., 2015; Vinyals et al., 2016). In recent years, this performance gap has dramatically narrowed due to advances in model architectures (e.g., prototypical networks, Fort, 2017; Snell et al., 2017) and learning paradigms (e.g., meta-learning, Finn et al., 2017).

However, a large performance divide persists for few-shot learning of relational or compositional categories, defined by abstract patterns of relations between multiple objects (see Figure 1). Unlike concepts for which the few-shot performance gap has narrowed, learning relational or compositional categories, like those in the Synthetic Visual Reasoning Test (SVRT), remains challenging even with extensive training on a vast number of examples (Fleuret et al., 2011). Moreover, in cases for which prolonged training does achieve high test accuracy, small visual perturbations of the objects dramatically reduce performance, suggesting that the learned concepts rely on non-relational visual features rather than genuine relational understanding (Kim et al., 2018; Messina et al., 2021; Puebla & Bowers, 2022).

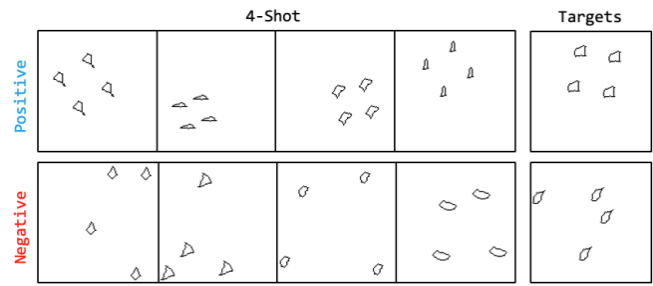

Figure 1: Example of a 4-shot compositional learning test from SVRT. The positive images display objects configured in a square, whereas the negatives randomly scatter the objects. Given both sets of images, the task is to classify a target image as positive or negative.

Meanwhile, cognitive psychologists have extensively studied how humans rapidly learn relational concepts. Decades of research have pinpointed two capacities that support this ability. First, evidence suggests that human representations of visual scenes are compositional: people perceive scenes as sets of objects structured by the relations between them with role information bound to each object (e.g., *inside (cat, box);* Hafri & Firestone, 2021; Wiesmann & Võ, 2023). This format ensures that *inside (cat, box)* is semantically distinct from *inside (box, cat)* and preserves role information that would be ambiguous in a flat list of objects and relations (e.g., *[cat, box, inside]*; Corral et al., 2018).

Second, humans acquire relational concepts from these compositional representations by abstracting their common relational structure. To identify shared structure, people compare compositional representations using analogical mapping, a process that aligns objects playing similar roles across images (Gick & Holyoak, 1983; Halford, Bain et al., 1998; Halford & Busby, 2007; Christie & Gentner, 2010; Kurtz et al., 2013; Jung & Hummel, 2015; Corral et al., 2018). For example, given *inside (cat, box)* and *inside (dog, house)*, we recognize that *cat* and *dog* fill the same role relative to their relations and therefore correspond to each other, rather than to *box* or *house*. When comparing across classes, mapping highlights critical relational differences (Winston, 1975; Gentner & Markman, 1994; Jee et al., 2013). This process has been formalized using graph-matching algorithms in which nodes represent objects and edges represent relations (Falkenhainer et al., 1989; Holyoak & Thagard, 1989; Lu et al., 2022).

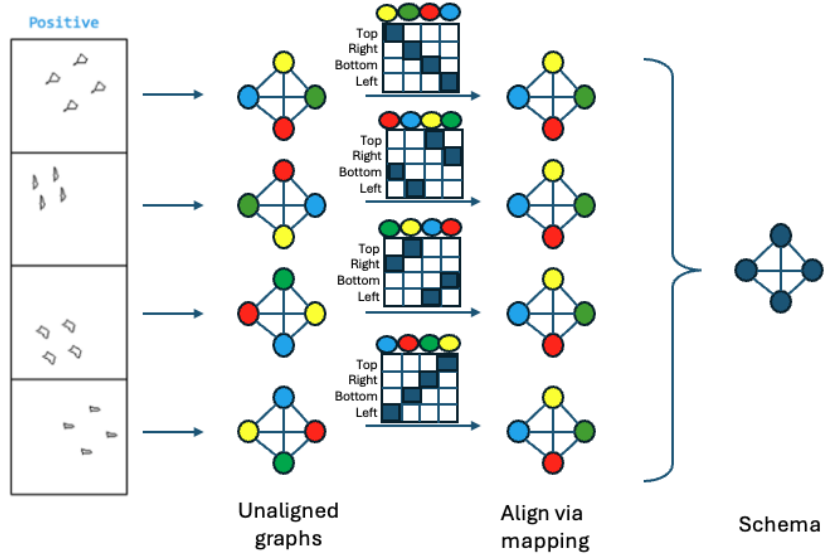

Figure 2: Diagram of schema formation in Probabilistic Schema Induction (PSI). Each exemplar image is first converted into a compositional representation in the form of a directed graph, with nodes representing objects and edges representing relations. PSI then aligns nodes and edges across exemplars to compute a prototype graph, or schema, by averaging over the aligned components. This alignment is discovered based on maximizing similarity between the schema and exemplar graphs.

**Probabilistic Schema Induction**

Here, we introduce Probabilistic Schema Induction (PSI), a model that combines deep learning with theories of human compositional concept learning and demonstrates human-like performance on few-shot visual concept learning of the SVRT. Our approach broadly mirrors that of classic prototype models of categorization: a class prototype is first formed from exemplar representations and then used to guide classification based on similarity to the target. As shown in Figure 2, our model is composed of several major parts. The first module aims to generate a compositional representation of each image in the form of a directed graph. The second module constructs a schema—the prototypical graph—for every class using PSI. The third module predicts the class of a target image by selecting the class whose schema is most similar to the target (also represented as a directed graph).

PSI departs from classic categorization models in three novel ways: representation, similarity, and prototypes. First, in prototype and exemplar models, every exemplar is represented as a *single* vector, where each value reflects the magnitude of a particular feature of the exemplar. In contrast, PSI decomposes an exemplar image into a graph consisting of objects and the relations between them. In this graph, each node is a vector capturing the visual properties of a single object. Each edge—a connection between two nodes—is a vector with values that capture the probability the two objects instantiate some set of visuospatial relations (e.g., *inside (A, B)* = 0.7, *sameShape (A, B)* = 0.2). As a result, a graph in PSI can be representationally more complex (or "heavier") than the single-vector format used in classic models. This increase in representational complexity—which imposes a greater cognitive load and may prompt a shift to single-vector representations under certain task conditions—is a core theoretical claim about how humans represent relational categories (Corral et al., 2018; Lee et al., 2023, in press).

Graph-based representations also satisfy three other theoretical properties of relational concepts. Specifically, they support *relational invariance*: a relation retains its identity across different entity pairs because it is represented separately from the entities themselves—for instance, as an edge distinct from the nodes it connects (Doumas & Hummel, 2005). Edges also enable *universal quantification*: because relations are encoded independently of particular entities, the same relation can apply to all relevant entities. Furthermore, directed graphs exhibit *role-filler independence*: the roles in a relation are encoded independently of the specific entities that fill them (Doumas & Hummel, 2005). To achieve this, every pair of nodes is connected by two directed edges, one in each direction. In one edge, a given node serves as the "sender" and the other as the "receiver"; in the reverse edge, these roles are swapped. Role identity is therefore determined not by explicit labels, but by edge direction. A similar format has been employed in analogical inference models by Lu et al. (2022) and Webb et al. (2023). Other models satisfy these constraints using, for instance, temporal synchrony (Hummel & Holyoak, 1997; Doumas et al., 2008, 2022).

The second and third ways in which PSI departs from classic categorization models are its novel formulations of similarity and prototypes. Classic models compute similarity as a function of the distance between two feature vectors, typically based on the difference in values of corresponding features. A class prototype is defined as the average feature vector across its instances. Critically, these definitions rely on the assumption that feature vectors share a fixed feature order: each feature occupies the same index across vectors for all exemplars (e.g., index $i$ always refers to *size*). This alignment in feature position ensures comparisons are always made between the same type of information. *Size* is never compared with *color* because vectors in the same space preserve the order of representational units. Distance is therefore a straightforward subtraction between feature vectors, and the prototype a straightforward mean.

However, unlike vector format, nodes and edges with the same indices across graphs do not necessarily share the same type of information. Standard verbal analogy tasks in a *A:B::C:?* format, by design, align roles by index (*A* shares the same role and relative index as *C*). However, alignment is not guaranteed in graphs parsed from visual data in which objects with corresponding roles across images may lie in different spatial positions. To appropriately compute similarity and prototypes without averaging across differing types of information, we propose using analogical mapping: each object in one graph maps to an object in another image based on similarity of objects and similarity of relations (e.g., Gentner, 1983; Holyoak & Thagard, 1989). Once an appropriate mapping is discovered (e.g., graph alignment), a prototype is computed as the average of nodes and edges with the same *alignment index*, and similarity is based on the average cosine similarity of only *aligned* nodes and the average cosine similarity of only *aligned* edges.

Although it would be ideal to extract object and relation information directly from visual input for compositional inference, most existing models can cleanly segment a scene into its constituent objects but still struggle to reliably infer the relations between objects. Given this limitation, we pose three questions. First, when provided with "ideal" graphs, composed of objects and relations directly sourced from the image generators of the SVRT (i.e., simulating perfect object segmentation and relation detection), how closely does PSI emulate human-like performance? Second, can PSI outperform a deep-learning variant of a prototype model that relies on singleton feature vectors—i.e., does the compositional structure of representations imposed by PSI enhance classification? Third, how does PSI perform when using pseudo-compositional but automatically generated representations that do not explicitly encode object-relation structure, such as patch embeddings from a pretrained vision transformer?

Here, we show that PSI is able to demonstrate human-like learning on the SVRT, outperform a singleton-vector prototype model, and perform better when using object-relation representations rather than pseudo-compositional patch embeddings. Our results suggest that structured, compositional representations confer an advantage in few-shot concept learning of compositional concepts. More broadly, these findings support the hypothesis that human visual reasoning relies not just on feature similarity, but on the ability to represent and compare structured representations. PSI thus provides a computational bridge between deep learning models and human concept learning.

## Model Architecture

**Mapping** As mentioned before, PSI addresses the challenge of invalid averaging across misaligned nodes and edges by first computing a mapping between nodes and between edges based on a combination of node similarity and edge similarity. This mapping identifies correspondences between nodes in an exemplar and nodes in the schema, thereby determining which nodes should be averaged across exemplars. PSI discovers a mapping between nodes by minimizing a loss function based on an overall graph-level similarity $G$ between a schema and its exemplars. Specifically, this loss function maximizes similarity $G$ based on two assumptions: (1) maximum similarity between a schema and its exemplars is achieved when mappings align similar object features and similar relation values, and (2) maximizing similarity between graphs approximates their "true" similarity.

Formally, a mapping between a schema and an exemplar is represented as a binary matrix, where rows correspond to schema nodes and columns to an exemplar's nodes. Each cell in this matrix is either 1 (mapped) or 0 (not mapped) with a one-to-one constraint: each row and column contains at most one "1," ensuring that each node is mapped to exactly one counterpart. A separate mapping matrix is learned between the schema and each exemplar. All matrix entries across exemplars are free parameters updated to minimize the loss function using backpropagation. Crucially, rather than updating a binary matrix itself, backpropagation (which provides continuous updates to parameters) updates mapping matrices that are continuously valued. At every step of backpropagation, the continuous matrices are converted into discrete permutation matrices using the Hungarian algorithm. These permutation matrices—binary and one-to-one—are used to compute the schema and the similarity between the schema and its exemplars. The loss based on this similarity is backpropagated to the underlying continuous matrices.

**Similarity** We define overall graph-level similarity $G$ as the weighted average of node-level similarity $G_{nodes}$ and edge-level similarity $G_{edges}$:

$$G = \alpha G_{nodes} + (1-\alpha) G_{edges}$$

Here, the alpha (α) parameters control the relative contribution of node and edge similarity: higher values place more weight on node similarity (i.e., object similarity) and lower values place more weight on edge similarity (i.e., relational similarity). Alpha is also treated as a parameter (along with the mapping matrices) and is constrained to a value between 0 and 1 through a sigmoid function at every step of backpropagation. We hypothesize that successful classification of compositional concepts—defined by a similar pattern of relations over objects—will be associated with lower final estimates of alpha, indicating a greater reliance on edge similarity.

We compute node similarity $G_{nodes}$ as the average cosine similarity between aligned schema-exemplar *node* pairs. Edge similarity $G_{edges}$ is computed as the average cosine similarity between aligned schema-exemplar *edge* pairs. Unlike node mappings, which are estimated, edge mappings are derived deterministically from the node mapping matrix. Specifically, an edge in the schema (from "sender" node *a* to "receiver" node *b*) is aligned with an edge in the exemplar (from "sender" node *c* to "receiver" node *d*) only if the "sender" nodes are mapped (*a*→*c* = 1) and the "receiver" nodes are mapped (*b*→*d* = 1). Then, the mapping value for the two edges is the product of the "sender" *a*→*c* mapping and the "receiver" *b*→*d* mapping. If either fails to map, the edge will not map either. Edges align only when their corresponding nodes align, reflecting the intuition that relations correspond only when their arguments do.

**Edge Weights and Contrastive Learning** Standard category learning tasks typically involve at least two categories. Learners must infer not only what defines a single category, but also what distinguishes it from others. To support this type of learning, we introduce an edge-weight vector that scales each relation within an edge, analogous to selective attention parameters in classic models (e.g., Nosofsky, 1986). The edge-weight vector is a free parameter constrained by a softmax function applied at each backpropagation step, ensuring that the weights are non-negative and sum to 1. Because PSI is trained to maximize similarity between a schema and its exemplars, the edge-weight vector gives PSI the flexibility to upweight relations that are consistent across exemplars of the same class. Due to the softmax constraint, increasing the weight of one relation necessarily reduces the relative influence of others, creating competitive pressure that helps suppress noisy, inconsistent relations. However, maximizing similarity within-classes alone does not guarantee that the relations that are upweighted are discriminative across classes. To encourage PSI to estimate the edge-weight vector so as to enhance consistent *and* discriminative relations, we incorporate contrastive learning terms in the loss function.

Specifically, when the task involves learning two categories, we compute node and edge similarity between the two schemas, denoted as $G_{nodes}^{C}$ and $G_{edges}^{C}$. These values are computed using an additional mapping matrix between the schemas (also a free parameter). In principle, we want to minimize edge similarity between schemas. However, doing so without constraints risks forming trivial node mappings (i.e., misaligning nodes to artificially reduce edge similarity). To avoid this, we simultaneously minimize edge similarity and maximize node similarity between schemas. Doing so forces PSI to simultaneously align schema nodes and search for a solution state that minimizes edge similarity under a non-trivial node mapping. Since schemas are computed from edge-weighed exemplars, one solution is to discover edge-weights that increase within-class similarity and reduce edge similarity. Relations that meet both criteria are likely to be consistent within-class and discriminative between-classes.

**Loss Function and Free Parameters** We define the total loss function as:

$$L = -G^P - G^N - G_{nodes}^C + G_{edges}^C$$

where P refers to the positive class and N refers to the negative class in a two-class setup. The first three terms are negated because the objective is to maximize similarity between a schema and its exemplars and maximize similarity between schemas' nodes (note that this models a special case where both positive and negative categories share similar object features). The final term is not negated because we aim to minimize edge similarity between schemas. In sum, the free parameters that are estimated by this loss function include the schema-exemplar mapping matrices, the schema-schema mapping matrix, the edge weights, and alpha. The mapping matrices are initialized as random values in the range [0, 0.01). We use an AdamW optimizer with learning rate 0.01.

**Classification** To classify a target image, we compute its similarity to each class schema using the final parameters obtained from backpropagation. During this stage, the schemas are fixed; we do not update either schemas or the edge-weight vector. Instead, we optimize new mapping matrices between the target and each schema. The contrastive loss terms are removed, and only schema-target similarities are maximized. The target's edges are augmented by the final estimated edge-weight vector, and similarity is computed using the final estimated alpha. The predicted class is the one with the schema with highest similarity to the target.

**Control Models** We compared PSI to two control models. The first is a single-vector prototype model that uses CLS tokens extracted from a pretrained vision transformer (DINOv2; Oquab et al., 2023). A CLS token is a vision transformer's single-vector representation of an image. Classification is based on the cosine similarity between the target image's CLS token and the average CLS token. The second is a pseudo-compositional variant of PSI, which uses DINOv2 patch embeddings as node representations. Vision transformers standardly process an image by dividing it into a grid of patches that then undergo refinement across layers of self-attention (Dosovitskiy et al., 2020). Patches inevitably cut across object boundaries, but the "cross-pollination" effect of self-attention in transformers may reconstruct object and relation information to some extent. This model uses no edges, edge weights, alpha, or contrastive loss; the loss function is based purely on node similarity of patches.

## Dataset

**SVRT** The Synthetic Visual Reasoning Test is an image generator that produces 23 compositional concepts, each consisting of positive and negative image classes (Fleuret et al., 2011). Each object in an SVRT image is a simple pixelated contour shape resembling an island (Figure 1). Objects vary in shape and size across both positive and negative images within a given problem (except when about sameness of shape). All negative examples in the SVRT are "hard" negatives: they share similar object properties as positives and differ only in relations over objects.

Problems span a wide range of relational rules. Roughly half are defined by first-order relations in which classes differ by a single pairwise relation. Examples include Problem 4, where a smaller object is inside a larger one (vs. outside the larger one); and Problem 16, where objects are mirrored along the vertical bisector (vs. not mirrored but located in same positions). The remaining half are defined by second-order relations in which classes differ by relations over pairwise relations. For example, Problem 10 is defined by four shapes organized into a square configuration (vs. random configuration; see Figure 1). Problem 12 is based on two small objects equally distant to a larger object (vs. unequally distant). The present version of PSI is only able to code first-order relations. At least one other model uses an analogy-style approach to the SVRT (Shurkova & Doumas, 2022).

**Forming Graphs** The SVRT image generator creates scenes by placing objects according to the structure of each problem. To extract object and relation representations, we modified the generator to save each object's pixel coordinates and compute pairwise relations between all objects. Using these coordinates, we reconstructed individual "object masks," or images that each show only a single object in its original spatial position. These masks were passed through a pretrained vision transformer (DINOv2), and the resulting CLS token was the feature vector for a node in the graph.

We defined seven relational features to describe pairwise relations: inside, touching, sameness of shape, normalized distance, mirrored, sameness of size, and reflection. All relations were binary (1 or 0) except for normalized distance, which was continuous. To introduce variability, we added Gaussian noise (mean = 0, sd = 0.1) to each relation value. For the continuous normalized distance value, noise was scaled to 20% of the random sample. Each edge in the graph was thus a 7-long vector coding these relations.

## Results and Discussion

We tested variants of PSI that either set alpha as a free parameter or fixed it to values of 1 (similarity is based entirely on node similarity), 0 (based entirely on edge similarity), or 0.5 (half node, half edge similarity). Figure 3 displays performance as a function of the number of few-shot examples on the SVRT, separately for first-order and second-order problems. To compare model and human performance, we computed root-mean-square error (RMSE) and mean absolute error (MAE) between each model's accuracy curve and the human curve, measured in percentage-point deviation across number of few-shot examples.

For first-order problems (Figure 3 left), PSI comes closest to reproducing human performance when alpha is a free parameter, deviating by 5.3% RMSE and 3.1% MAE. Interestingly, PSI with patches follows second with 6.7% RMSE and 5.7% MAE. The CLS model has slightly worse fit with 8.1% RMSE and 7.7% MAE, and also performs worse than PSI with adaptive alpha ($\beta$ = -0.56, $z$ = 7.95, $p$ < .0001) but similarly as PSI patches ($\beta$ = 0.10, $z$ = 1.54, $p$ 0.12). The worst matches to human performance are PSI with fixed alpha (RMSE ≥ 11.4%, MAE ≥ 10.8%). For second-order problems (Figure 3 right), no model captures human performance adequately; the closest fit is PSI with patches (RMSE = 12.3%, MAE = 11.7%).

Among variants of PSI, first-order problem performance is highest when similarity is based entirely on edge similarity (compared to adaptive alpha: $\beta$ = 0.78, $z$ = 8.82, $p$ < 0.0001).

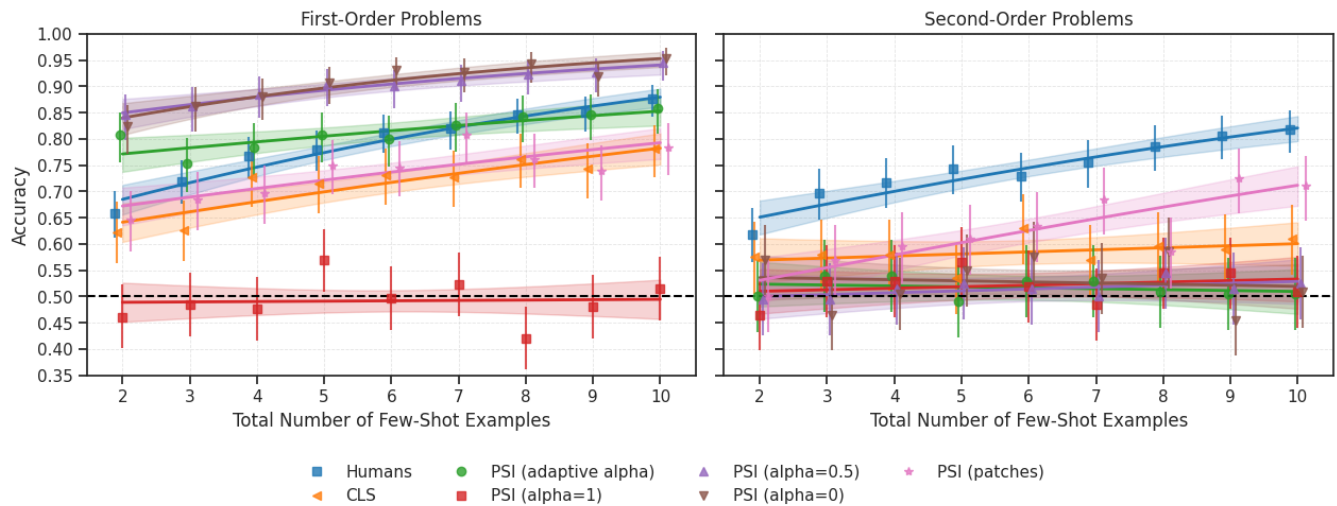

Figure 3: Accuracy by total number of few-shot examples, separately for first-order and second-order problems. Curves are logistic regressions. Error bars/bands are binomial 95% confidence intervals. Dashed lines indicate chance.

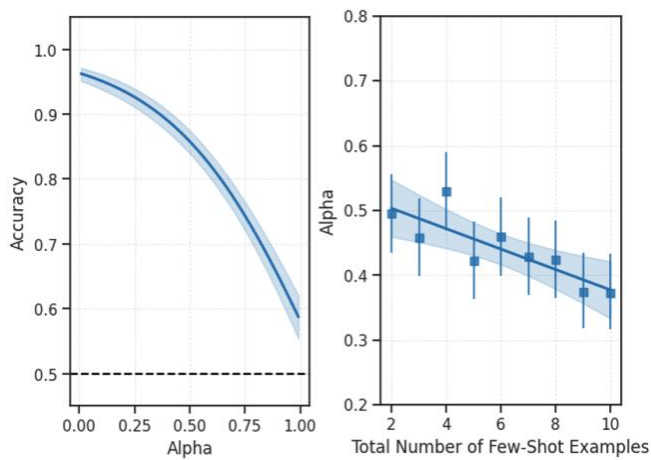

Figure 4. Left: Accuracy increases when alpha decreases, emphasizing relational similarity. Right: Alpha decreases as PSI is provided with more examples, suggesting that relational similarity grows stronger with more examples. Curves reflect fitted logistic regressions.

In contrast, relying solely on node similarity yields performance statistically indistinguishable from chance ($z = -0.79$, $p = 0.43$). Interestingly, combining node and edge similarity (alpha of 0.5) performs no differently than using edge similarity alone ($\beta = -0.054$, $z = -0.54$, $p = 0.59$). These results are consistent with the hypothesis that success on compositional concept classification depends more on similarity of relations between objects than similarity of object-level features.

Consistent with these results, adaptive alpha decreases as accuracy increases ($\beta = -2.94$, $z = -17.3$, $p < 0.0001$; Figure 4 left) and as more few-shot examples are provided ($\beta = -0.064$, $z = -3.96$, $p < 0.0001$; Figure 4 right). However, on average, alpha converges on values slightly below 0.5 (mean = 0.44, sd = 0.43), reflecting a mixture of contributions from relational and object similarity. The fact that alpha does not fully collapse to 0 (full edge similarity) suggests that human learning may rely in part on object-level similarity.

In addition, PSI with adaptive alpha progressively increases the weight assigned to class-distinguishing relations as more few-shot examples are provided, while downweighing non-distinguishing relations (Figure 5 left). These shifts affect accuracy: higher weights on distinguishing relations yield higher accuracy, while higher weights on non-distinguishing relations yield lower accuracy (Figure 5 right). Together, these results suggest that PSI's human-like learning curve arises from a dual adaptation mechanism: (1) a shift in alpha toward relational similarity, and (2) selective weighting of relevant relations.

## Conclusion

We introduced Probabilistic Schema Induction (PSI), a prototype model of compositional concept learning that departs from classic approaches to category learning by using compositional representations and a novel formulation of similarity and prototypes based on analogical mapping. We showed that PSI not only achieves human-like performance, but also outperforms deep-learning baselines on a few-shot compositional classification task. Critically, PSI's success stems from its ability to adaptively emphasize relational similarity and to upweight relations that distinguish classes. Interestingly, the model does not fully converge on purely relational similarity, suggesting that humans may rely on a blend of object similarity and relational similarity.

The version of PSI presented here is able to handle first-order relation problems by representing pairwise relations between objects as edges. However, it does not represent relations between pairwise relations and is therefore unable to capture second-order relational similarities. In principle, second-order relations can be modeled as the difference between pairwise relations, coupled with an adaptive alpha that weights three tiers of abstraction: node, first-order edge, and second-order edge similarity. In addition to extending the model to handle second-order relations, future work should explore whether PSI's alpha values can predict human accuracy in far generalization tests, and develop methods that automatically parse object and relation representations directly from raw visual input.

## Acknowledgements

We thank research assistants Rui Yu and Anna Wang, as well as members of the UCLA Reasoning Lab and UCLA Computational Vision and Learning Lab. This work is supported by AFRL S03032-01-0-LSC. This paper was approved for public release under case AFRL-2025-1067 and represents the views of the authors and does not represent any position of the United States government, US Air Force, or Air Force Research Laboratory.

## References

Christie, S., & Gentner, D. (2010). Where hypotheses come from: Learning new relations by structural

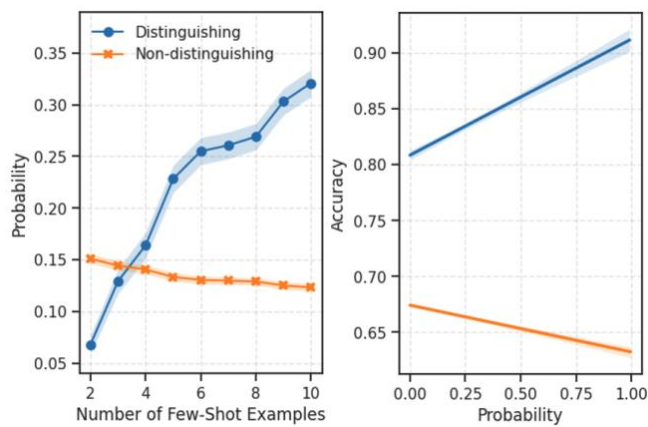

Figure 5. Edge weight (probability) by number of few-shot examples (left) and accuracy by edge weight (right), separately for class-distinguishing and non-distinguishing relations of a problem. Lines on the right reflect linear regressions. Error bands are 95% confidence intervals.